# Comparing Simulation Output Accuracy of Discrete Event and Agent Based Models: A Quantitative Approach

**Mazlina Abdul Majid**
School of Computer Science
University of Nottingham,
Nottingham, UK
+44 (0)115 8465526
mva@cs.nott.ac.uk

**Uwe Aickelin**
School of Computer Science
University of Nottingham,
Nottingham, UK
+44 (0)115 9514215
uxa@cs.nott.ac.uk

**Peer-Olaf Siebers** School of
Computer Science
University of Nottingham,
Nottingham, UK
+44 (0)115 8465526
pos@cs.nott.ac.uk

**Abstract**

In our research we investigate the output accuracy of discrete event simulation models and agent based simulation models when studying human centric complex systems. In this paper we focus on human reactive behaviour as it is possible in both modelling approaches to implement human reactive behaviour in the model by using standard methods. As a case study we have chosen the retail sector, and here in particular the operations of the fitting room in the women wear department of a large UK department store. In our case study we looked at ways of determining the efficiency of implementing new management policies for the fitting room operation through modelling the reactive behaviour of staff and customers of the department. First, we have carried out a validation experiment in which we compared the results from our models to the performance of the real system. This experiment also allowed us to establish differences in output accuracy between the two modelling methods. In a second step a multi-scenario experiment was carried out to study the behaviour of the models when they are used for the purpose of operational improvement. Overall we have found that for our case study example both, discrete event simulation and agent based simulation have the same potential to support the investigation into the efficiency of implementing new management policies.

## 1. INTRODUCTION

Simulation has become a preferred tool in Operation Research for modelling complex systems. Studies in human behaviour modelling have received increased focus and attention from simulation research in the UK [Robinson 2004]. The research in human behaviour modelling has been applied to various application areas such as manufacturing (e.g. [Siebers 2004]), healthcare (e.g. [Brailsford et al. 2006]), military operations (e.g. [Wray and Laird 2003]) crowd behaviour (e.g. [Loftin et al. 2005]) retail management (e.g. [Siebers et al. 2008]) and consumer behaviour (e.g. [Schenk et al. 2007]). As found in the literature, some researchers choose Discrete Event Simulation (DES) as a means to investigate their human behaviour problems; others choose Agent Based Simulation (ABS) for this purpose. The choice of which simulation method to use relies on the individual judgment of the modeller and their experienced with the modelling method. The issue here will be how accurate and difference the simulation output will be when we model human behaviour using both DES and ABS model. The representation of human behaviour contains complexity and variability; therefore when investigating such systems it is very important to choose a suitable technique. Consequently, we have done some quantitative experiments and our findings will be discussed in this paper.

In this research we aim to provide an empirical study in order to find out more about the differences in output accuracy by comparing traditional DES and ABS. The main difference between traditional DES and ABS is that in traditional DES the modelling focus is on the process flow while in the ABS the modelling focus is on the individual entities in the system and their interactions. To achieve our aim we will compare the simulation output accuracy of DES and ABS models when modelling human reactive behaviour in a department store. In this context human reactive behaviour means responding to a request. For example, a sales staff member provides help when needed. Statistical tests will be used to establish if the differences in output accuracy of the different modelling methods are significant.

The remaining content of this paper is as follows: In Section 2 we explore the theory and characteristics of the three major OR simulation methods - DES, ABS, and System Dynamics (SD). Here we also discuss comparisons between the different simulation methods we have found in the literature. Section 3 describes our case study planning, field work and our model design. In Section 4 we describe our experimental setup and then analyse and discuss our results. We have conducted a validation and a multi-scenario experiment to test our models' behaviour under real world conditions. Finally, in Section 5 we draw some conclusions and summarise the current progress.

## 2. LITERATURE REVIEW

At present, several tools and techniques can be used to model a system. Over the last three decades, simulation has become a frequently used modelling tool in Operation Research (OR) [Kelton 2007]. Its ability to support studies of complex systems has made simulation the preferred choice of academics and practitioners in comparison to analytical approaches. The simulation modelling paradigms used in OR can be classified in three groups: (traditional) discrete event modelling, system dynamics modelling, and agent based modelling.

DES models represent a system based on chronological sequences of events where each event changes the system state in discrete time. SD models represent real world phenomena using stock and flow diagrams, causal loop diagrams and differential equations. In contrast, ABS models comprise of a number of autonomous, responsive and proactive agents which interact with each other to achieve their objectives.

When looking for existing comparisons between DES, SD and ABS models we found some papers about this topic in the literature; however none of them was focusing on modelling human behaviour. Some of the relevant papers comparing simulation technique with regards to model characteristics are listed in Table 1.

A review of existing comparisons between SD and DES is presented by [Tako and Robinson 2006]. They have reviewed 65 journal articles from 1996-2006 that compare model building, philosophies and model use of SD and DES models. They concluded that in most areas (manufacturing, supply chain management, etc.) SD has been used for the strategic planning while DES for the operational planning.

Existing comparison of DES and ABS is presented by [Pugh 2006] and [Yu et al. 2007]. Pugh states by looking into the model characteristics that DES and ABS models both represent M/M/1 queuing systems well but he found that ABS models are much more difficult to construct compared to DES models. [Yu et al. 2007] have conducted a quantitative comparison between DES and ABS model characteristics in the field of transportation. They found that DES model appears to have greater value in the simulation software internal properties such as by building DES models in their simulation software required more model blocks where as ABS models required less classes. This shows that even though DES and ABS can both model the system under investigation, but their approach are different [Becker 2006].

We found just one literature that looked at all three modelling techniques [Owen et al. 2008]. They looked into establishing a framework for comparing the different modelling techniques, stating that a framework is essential in assisting the supply chain practitioners by matching their modelling problem with a suitable modelling paradigm.

Table 1: Existing study in comparing simulation technique regarding model characteristics.

| Techniques | Researchers; Area | Findings |
|---|---|---|
| SD and ABS | Wakeland et al. 2004; Biomedical | Found that the understanding of the aggregate behaviour in SD model and state changes in individual entities in ABS model is relevant in the study. |
| SD and DES | Morecroft and Robinsion 2006; Fisheries | Found that SD and DES are implementing different approaches for modelling but both are suitable for modelling systems over time. |
| DES and ABS | Becker et al. 2006; Transportation | Found that DES is less flexible than ABS; it is difficult to model different behaviours of shippers in DES. |

As conclusion to findings presented here, we can say that all of them agree that choosing the right modelling technique is essential in ensuring a good representation of the selected problem in the different areas.

We found a disparity in the quantity of work comparing SD and ABS or SD and DES to that comparing DES and ABS (which was very little) and we have found no work on comparing the accuracy of DES and ABS results for the study of human centric systems. We have chosen to fill this gap and focus our efforts on comparing DES and ABS models of human centric systems with regards to their output accuracy compared to the real system modelled.

To study the differences between ABS and DES models we will look at management practices and their influences on staff performance and customer behaviour in the retail business. Related research has mainly focused on consumer behaviour (e.g. [Schenk et al. 2007]). However, research in management practices has started to evolve as described by [Siebers 2007; 2008]. As we discussed above, much work has been done in comparing simulation techniques in the field of transportation and supply chain management and researchers have focused on model characteristics.

## 3. CASE STUDY FIELDSWORK

In order to achieve our aim we have used a case study approach. The research has focused on the operation of the main fitting room in the womenswear department of one of the top ten retailers in the UK (see Figure 1). For a client the goal of such a simulation study could be to identify the potential impact for fitting room performance when having different numbers of sales staff permanently present in the fitting room.

We have studied reactive staff behaviour which relates to staff responding to the customer when being available and requested. Based on our case study observations we have developed some conceptual models. Our DES conceptual model (see Figure 2) is represented by a flow chart diagram as in DES we focus on process flow. Our ABS conceptual models (see Figure 3, 4 and 5) are state chart diagrams for the different types of agents we have to represent (in our case customers, staff and fitting room) as in ABS we focus on the individual 'actors' and their interactions.

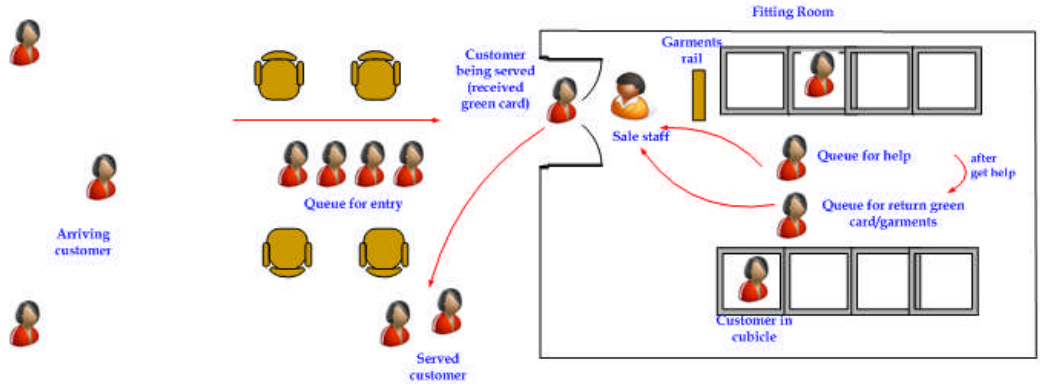

Figure 1: An illustration of the main fitting room operation

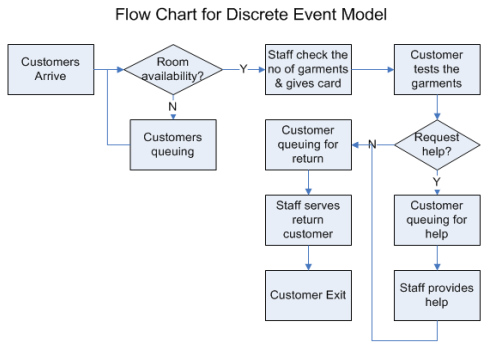

Figure 2: Flow chart for DES model

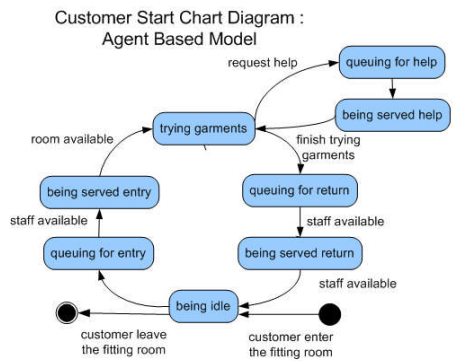

Figure 3: State chart for ABS model (customer)

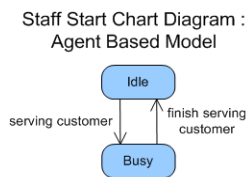

Figure 4: State chart for ABS model (staff)

In the fitting room operation, the staff have to do three distinct jobs: (1) counting the number of garments and handing out the fitting room's card (contains number of clothes and room number) when the customer enters the fitting room area, (2) providing help while customers are in the fitting room, and (3) taking back the fitting room's card and any unwanted garments when the customer is leaving the fitting room area.

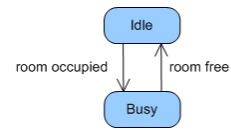

Figure 5: State chart for ABS model (fitting room)

## 4. EXPERIMENTATION

Two similar simulation models were developed from the conceptual models presented in Section 3 using the multi-paradigm simulation software AnyLogic™ [XJ Technologies] one implemented as a DES model and one implemented as an ABS model. Both models were constructed as conventional M/M/1 queuing systems. They consist of an arrival process (customers), three single queues (customer entry queue, customer return queue, customer help queue), and resource (sales staff).

The arrival process we observed in the real system over the cycle of a typical day is shown in Figure 6. In our simulation models we have modelled the arrival process using an exponential distribution with an annual changing arrival rate in accordance with the arrival rates shown in Figure 6.

Both simulation models use the same model inputs. Therefore, if we see any differences in the model outputs they will be due to the differences between the two modelling techniques. The simulation models terminate after a business day (8 hours), mimicking the operation of the real department store. We conducted 100 replications for each set of parameters.

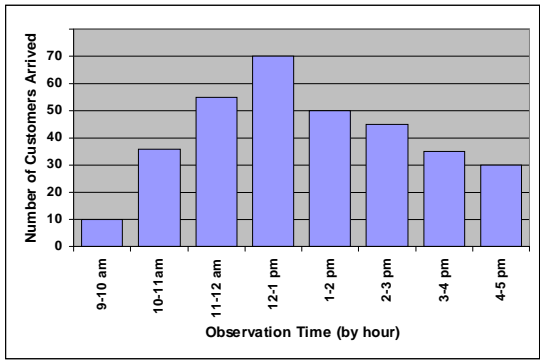

Figure 6: Distribution of customer arrival in the real system on a typical day
.
### 4.1. Model Validation

For our validation experiment we have used black box validation where we compared the simulation outputs from DES and ABS with the real system output in terms of quantities. We have one member of staff that does all three jobs mentioned above, job 1 (counting garments on entry), job 2 (providing help) and job 3 (counting garments on exit).

We have defined some hypotheses we wanted to test during our validation experiment. Here we have only stated the null hypotheses, assuming that the alternative hypotheses are always the opposite of null hypothesis. The two main hypotheses for the validation experiment are:
- $Ho_A$ = Our DES model is a good representation of the real system
- $Ho_B$ = Our ABS model is a good representation of the real system

As our comparative measure for judging the goodness of our representation we have chosen the *mean waiting time* from the three queues. This was the only performance data we were able to collect from the real system. For testing if the collected data is normally distributed (which is important for choosing the correct statistical analysis method) we used an informal approach, comparing a histogram of the residuals of the collected data to a normal probability curve. The comparison indicated that our data is probably not normally distributed, so we need to use non-parametric tests when analysing our data based on this performance measure.

#### 4.1.1. Comparing Medians Using a Non Parametric Test

If data is normally distributed the measures of central tendency (e.g. mean, median and mode) are the same since the normal distribution is symmetric. However, as our data is not normally distributed, the mean and median will have different values and to compare the median values we have chosen the non parametric Mann-Whitney statistical test, to confirm or disconfirm the following hypotheses:
- $Ho_C$ = Average customer waiting times resulting from our DES model are not significantly different to the ones observed in the real system.
- $Ho_D$ = Average customer waiting times resulting from our ABS model are not significantly different to the ones observed in the real system.

For performing the Mann-Whitney test we have used the open source statistical software package R [The R Foundation for Statistic Computing]. The median of the waiting times from DES and ABS models and the real system were calculated for this test. We have chosen 0.05 as our significance level. A test result (p-value) higher than 0.05 will allow us to accept a null hypothesis, otherwise we have to reject it. Testing our DES model results against the real system measures reveals a p-value of 0.3269. Testing our ABS model results against the real system measures reveals a p-value of 0.2958. Since both p-values are above our chosen level of significance (0.05) we fail to reject our two hypotheses $Ho_C$ and $Ho_D$. In addition, the fact that the p-value for our ABS model is slightly smaller compared to the p-value of our DES model conforms to our findings from Section 4.1.1.

From our statistical test results of the measures of central tendency we can confirm that the average customer waiting times resulting from both simulation models (DES and ABS) are not significantly different to the ones observed in the real system. Next we look at the variability of our performance measure (waiting time) to see if the variability we get from our simulation models matches the variability we can observe in the real system.

#### 4.1.2. Comparing Output Variability

In this test, we compare the variability of the results from the simulation runs with the performance variability occurring in the real system. This is done in two steps. First, we look at frequency plots of customer waiting times for a single day (which equates to a single simulation run). Here we use the results of a single day and single simulation run as we only have the complete real system observation data for a single day. In a second step, we calculate the variance (a measure of dispersion) of customer waiting times. This allows us to study the spread of the residuals from the mean values of our performance measure (customer waiting time) for the simulation models and the real system and therefore to compare the variability of the output data on a statistical basis. The two hypotheses we are testing are as follows:
- $Ho_E$ = Average customer waiting times resulting from our DES model show similar variability compared to those observed in the real system

- Ho$_F$ = Average customer waiting times resulting from our ABS model show similar variability compared to those observed in the real system

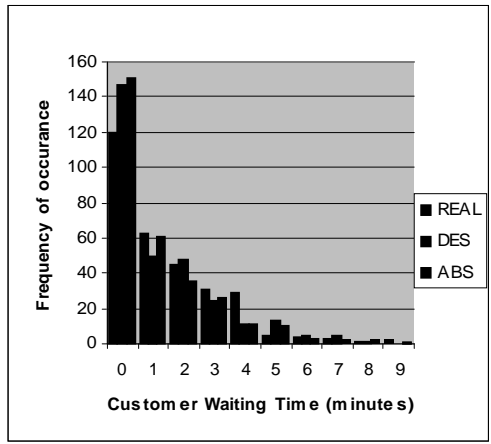

Figure 7 : Frequency distributions for customer waiting times

The frequency distribution histogram for the DES and ABS models and the real system is shown in Figure 7. They show the typical shape of exponential distributions. We can see that in the real system people are queuing slightly more frequently (customer waiting time > 0) compared to DES and ABS results. It can be observed that the gradient for the real world distribution is lower compared to the DES/ABS distributions. This holds up to a customer waiting time of 4 minutes. After this, there is a big change in the real system distribution which is not present in the DES/ABS distributions. This seems to be some natural or artificial boundary in the real system that is not reflected in our simulation models (maybe it reflects the maximum time that people are willing to wait in a queue).

In Section 4.1.1, we found that the mean customer waiting time appearing in the real system and the simulation results are not significantly different. However, mean values do not tell the whole story about the system. One also has to consider variability to get a complete picture about differences between the results. One way to do this is to calculate the variance. Measuring the variability (e.g. variance) of the model output is also useful to validate the models we use. Variability exists in the simulation output due to the randomness in the simulation input such as probability distributions. We can measure the variability is by estimating the variance. The variance shows how close the simulation outputs i.e. waiting time, in the distribution are to the middle of the distribution. It defines as the average squared difference of the outputs from the mean. [Lane 2003].

Table 2: Descriptive statistics for validation experiment (including variance)

| Models | Mean waiting time (minutes) | Standard deviation | Variance |
|---|---|---|---|
| Real system | 1.68 | 1.73 | 3.01 |
| DES | 1.69 | 1.59 | 1.96 |
| ABS | 1.61 | 1.70 | 2.89 |

Looking at Table 2 we can see that the variance of ABS compared to the real system is similar (4% difference) while variance of DES to the real system significantly different (35%). This gives us a different result of what we have observed in our first test where we failed to reject our hypotheses looking just at the mean value and the standard deviation (see Section 4.1.1). The reason for the differences in variance between DES and ABS is probably due to queuing discipline inherited in DES models where the system is 'over' organised, compared to ABS models where organisational structures are more related to the real world (perfectly organised queues are artificial construct). For this test then we fail to reject our hypothesis Ho$_E$ which states that the average customer waiting times resulting from our ABS model show similar variability compared to those observed in the real system, but we have to reject our hypothesis Ho$_F$ which states the dissimilarity for our DES model.

### 4.1.3. Validation Experiments Conclusions

In conclusion, we can say that we have found no significant statistical differences regarding the output accuracy of DES models and ABS models when simulating reactive behaviour. The same holds when comparing our simulation model outputs with the real system performance. This is what we would have expected, as the real system we have modelled is a typical queuing system and we have not added any features that would be unique to one of the simulation methods. The only difference we found was that the level of variance in a DES model was significantly lower compared to ABS and real system variance. We need to keep this in mind if we employ DES to simulate the behaviour of human centric systems.

However, overall we conclude that both simulation models (independent from the simulation method used) are a good representation of the real system, when focusing on reactive behaviour. Therefore, we fail to reject our hypotheses Ho$_A$ and Ho$_B$. What needs to be tested in the future is if we fail to reject the hypotheses Ho$_A$ and Ho$_B$ as well when we model proactive behaviour.

### 4.2. Comparing Multi-Scenarios

In this section, we look at different management policy scenarios regarding the staffing levels of the fitting room. These kinds of scenarios could be used to study how one could improve the performance of the fitting room operation that we have modelled in our validation experiment. We will use the results of our validation experiment as Scenario 1 and will measure the improvement in system performance against it. The purpose of this experiment is to see how both simulation methods behave when testing different scenarios. Also, we will keep an eye on the actual performance predicted by the simulation models. We have formulated the following two hypotheses:

- $Ho_G$ = Both simulation methods produce similar results when used in a multi-scenarios investigation
- $Ho_H$ = When allocating more staff to the fitting room our simulation model shows a reduction in customer waiting times, and time spend in the system.

Table 3 : Model Scenarios

| Scenario | Reactive Behaviour | | |
|---|---|---|---|
| | Job 1 (Count garments; give | Job 2 (Provide help) | Job 3 (Take unwanted garments; take card) |
| 1 | Staff 1 | Staff 1 | Staff 1 |
| 2 | Staff 1 | Staff 2 | Staff 1 |
| 3 | Staff 1 | Staff 2 | Staff 3 |

Overall, we have tested the efficiency of three different management policies with varying number of sales staff. The specific setup for these scenarios can be found in Table 3. In addition to the simulation setup, only 10% of the arriving customers will request help and all others will only return the fitting room's card before leaving. The simulation model outputs and their statistics are shown in Table 4.

For analysing the multi-scenario output of our experiments a paired-t CI test together with the Bonferroni inequality procedures were used. The paired-t CI test helps to test the statistical significance of differences between the output of two scenarios by comparing their CIs. The Bonferroni inequality procedure is an extended paired-t confidence method for comparing more than two scenarios.

A more detailed discussion about both methods can be found in [Clark and Yang 1986]. Instead of comparing CIs, there are other statistical methods such as the ranking and selection method [Law 2007] and many other statistical procedures for finding the best scenario (see [Goldsman and Nelson 2001] for a list). We have chosen a paired-t CI test together with the Bonferroni inequality procedure because these are complementary to each other and excellent tools for selecting the best scenarios [Swisher et al. 2003].

For our experimental analysis, we have chosen to compare the means of *customer waiting time* and *customer time in system*. The results from the simulation model setup used for the validation experiments have been chosen as the reference point for the analysis. We have selected 0.05 as our significance level. The Bonferroni inequality procedure states that if we want to make c confidence interval statements (c = s - 1, where s is the number of scenarios) with an overall significance level of α, the individual confidence interval should be formed with a significance level of α / c. Therefore, in our case the scenario significance level is 0.05 / 2 = 0.025. We have conducted 100 replications (for the DES and the ABS model, respectively) for each scenario.

To look into the relationship of the test result, we have followed the paired-t CI rules as discussed in [Robinson 2004] and shown in Figure 8 below. Based on these rules, we have produced our comparison results. These are shown in Table 5.

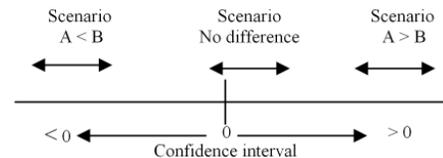

Figure 8: The relationship of the result in a paired-t CI

The results show that there is no significant statistical difference between Scenario 1 (S1) and Scenario 2 (S2) outputs between the DES and the ABS model. We believe this is because the difference in the number of customers who request help from the sale staff is very small (only 10% of customers). For the comparison of S1 with Scenario 3 (S3) the case is different. For both simulation methods and both output measures, the analysis shows a significant statistical difference between the output values. This can be explained by the division of work between staff. In S1 one staff member does all three jobs meanwhile in S3 there are three members of staff each doing one job.

Table 4: Simulation outputs for Scenarios in DES and ABS models

| Scenario | Performance measure (minutes) | DES model | | | | ABS model | | | |
|---|---|---|---|---|---|---|---|---|---|
| | | Mean | SD | 95% CI | | Mean | SD | 95% CI | |
| | | | | Lower limit | Upper limit | | | Lower limit | Upper limit |
| 1 | Waiting time | 1.69 | 1.59 | 1.37 | 2.00 | 1.61 | 1.70 | 1.27 | 1.94 |
| | Time in system | 8.79 | 0.98 | 8.60 | 8.99 | 8.48 | 1.64 | 8.14 | 8.80 |
| 2 | Waiting time | 1.52 | 1.01 | 1.32 | 1.72 | 1.45 | 1.61 | 1.13 | 1.77 |
| | Time in system | 8.58 | 0.67 | 8.45 | 8.71 | 8.37 | 1.87 | 8.00 | 8.74 |
| 3 | Waiting time | 0.89 | 0.70 | 0.75 | 1.03 | 0.80 | 1.08 | 0.59 | 1.01 |
| | Time in system | 8.10 | 0.34 | 8.03 | 8.17 | 7.48 | 1.94 | 7.10 | 7.87 |

Table 5: Results from the paired-t CI test and Bonferroni inequality procedure.

| Model | Performance measure | Scenario compariso | 97.5% CI for differences | Conclusion |
|---|---|---|---|---|
| DES | Waiting time | Scenario 1 to 2 | 0.33 | No difference |
| | | Scenario 1 to 3 | 0.42 +1.17 | S1> S3 |
| | Time in system | Scenario 1 to 2 | 0.43 | No difference |
| | | Scenario 1 to 3 | 0.47 + 0.93 | S1> S3 |
| ABS | Waiting time | Scenario 1 to 2 | 0.33 | No difference |
| | | Scenario 1 to 3 | 0.33 + 1.29 | S1> S3 |
| | Time in system | Scenario 1 to 2 | 0.21 | No difference |
| | | Scenario 1 to 3 | 0.41 + 1.58 | S1> S3 |

When we consider the results presented in Table 4 and Table 5 together we can make the following recommendations. When looking at the mean value of customer waiting times (Table 4), S3 has the smallest customer waiting times in the fitting room operation (for both simulation methods) and the paired-t CI in conjunction with the Bonferroni inequality procedure confirms that S3 is the best solution (Table 5). However, if some constraints exist, e.g. the costs for employing more staff, S1 would be the best solution as there is no significant difference between the performance of S1 and S2; S1 still produces small waiting times with an average of less than 2 minutes.

In conclusion, for this experiment we fail to reject both our hypotheses. Therefore, we can say that both simulation methods (DES and ABS) produced similar results when used in a multi-scenarios investigation. Furthermore, we found that when allocating more staff to the fitting room our simulation model will show a reduction in customer waiting times, and time spends in the system.

## 5. CONCLUSION AND FUTURE WORK

In this paper, we have investigated which simulation method (DES or ABS) is suited to create a good representation of the human centric system when considering only human reactive behaviour. In this instance 'good' means that the simulation mod model output matches the performance observed in the real system. This does not only include the measures of central tendency, but also measures of dispersion.

For our test, we have conducted a case study where we have analysed the fitting room operation of a womenswear department. Once we had a good understanding of the operation of the system, we were able to build some simulation models using different modelling techniques (DES and ABS). Then we conducted two experiments to compare the behaviour and outputs of our simulation models to those observed and measured in the real system. The first experiment was a validation experiment while our second experiment was a multi-scenario comparison.

In our experiments, we were able to demonstrate that DES and ABS models are equally good representations of the real system when we are interested in modelling human reactive behaviour. This only holds when we look at measures of central tendency. When looking at variability, DES does not reflect the true variability comprised in the real system in an appropriate way. However, the big advantage of DES is that it is more commonly used, and it is much more used, in particular in industry. Moreover, model design seems to be easier. Another advantage is the DES can be more easily validated than ABS, as you model individual entities in ABS only and not the macro behaviour often a validation on the macro level is not possible. Once this limitation of DES is known one can consider it when drawing conclusions from the simulation study results.

A problem we had with our current study is our choice of our performance measure. Customer waiting time is not a very robust measure of system performance, as it turned out that customers in the department store that we used for our case study did not have to queue very often. To make our investigation more robust we plan to repeat the experiments using a second performance measure; staff utilisation. However, before we can do that we will have to get access again to the case study department to collect the data from the real system. In addition, we plan to investigate more scenarios such as staff doing multiple jobs and staff changing the serving order.

So far, we have looked at scenario that we assumed you could model by using any of the two modelling techniques and you would get very similar results. In the future we want to look at aspects of human behaviour you would typically only model in ABS, for example proactive behaviour. Here the research question we want to answer would be: Is it worthwhile to put additional effort in to model these kinds of features in an OR simulation study, or do they not have a big impact on the conclusions you can draw from your simulation study?

**AUTHORS BIOGRAPHIES**

**MAZLINA ABDUL MAJID** is a PhD student in the School of Computer Science, University of Nottingham. Her interest is in discrete event simulation and agent based simulation. Her email is <mva@cs.nott.ac.uk>.

**UWE AICKELIN** is a Professor in the School of Computer Science, University of Nottingham. His interests include agent based simulation, heuristics optimisation, artificial immune system. His email is <uxa@cs.nott.ac.uk>.

**PEER –OLAF SIEBERS** is a Research Fellow in the School of Computer Science, University of Nottingham. His main interest includes agent based simulation and human complex adaptive system. His email is <pos@cs.nott.ac.uk>.